# On the power spectral density applied to the analysis of old canvases

Francisco J. Simois*, Juan José Murillo Fuentes

*Department of Signal Processing and Communications, Escuela Técnica Superior de Ingeniería, Universidad de Sevilla, Camino de los Descubrimientos s/n, 41092, Seville, Spain.*

*Corresponding author. Tel.: + 34 954 48 81 33; fax: +34 954 48 81 33.

E-mail addresses: fjsimois@us.es (Francisco J. Simois), murillo@us.es (Juan José Murillo Fuentes).

**Abstract**

A routine task for art historians is painting diagnostics, such as dating or attribution. Signal processing of the X-ray image of a canvas provides useful information about its fabric. However, previous methods may fail when very old and deteriorated artworks or simply canvases of small size are studied. We present a new framework to analyze and further characterize the paintings from their radiographs. First, we start from a general analysis of lattices and provide new unifying results about the theoretical spectra of weaves. Then, we use these results to infer the main structure of the fabric, like the type of weave and the thread densities. We propose a practical estimation of these theoretical results from paintings with the averaged power spectral density (PSD), which provides a more robust tool. Furthermore, we found that the PSD provides a fingerprint that characterizes the whole canvas. We search and discuss some distinctive features we may find in that fingerprint. We apply these results to several masterpieces of the 17th and 18th centuries from the Museo Nacional del Prado to show that this approach yields accurate results in thread counting and is very useful for paintings comparison, even in situations where previous methods fail.

## 1. Introduction

Lately there has been a wider interest on acquiring and processing artwork image data, thanks to the greater computing capabilities and digital storage. Advances in technology for image data acquisition and the wide range of imaging modalities currently available have made that museums assemble vast digital libraries of images of their collections, not only for archiving but also for analyzing their pieces of art [1]. Moreover, an increasing number of scientists have approached this field to apply digital image processing and data analysis techniques. These techniques have been proved useful for many crucial issues in artwork diagnostics, such as assessment of the conservation state, knowledge of the realization techniques, evaluation of the historical period and attribution of the painting, and the possibility of keeping trace of any modification of the artwork shape [2]. Therefore, interdisciplinary interaction between art historians and signal processing researchers is of greatest interest to elaborate image analysis tools that help to make repeatable automatic procedures and diagnostics on paintings [3]-[5].

Traditionally, one of the most common canvas analyses is based on X-ray images of paintings. The canvas fabric is initially coated with a priming substance, which generally contains lead. This ground layer varies in thickness according to the weave threads of the canvas, affecting the X-ray absorption. This makes the fabric pattern be showed in radiographies of the painting [6]. Until a few years ago, art experts used X-rays to analyze only the composition of the visible and hidden paint layers on the canvas. More recently, however, researchers have realized that the threads pattern may carry also important art-historical information [7].

The first pattern measurement to be used was the thread densities [8], [9]. These densities are not specific to each artwork, but characterize the roll from which the canvas was cut. Specifically, the weaving process itself results in the thread densities showing striped arrangements, in the direction of both warp (corresponding to the length of the roll placed on the loom) and weft (corresponding to the width of the roll) [10]. Thus, matching of the densities maps of two paintings makes it possible to identify aligned pieces of canvas from the same original roll, which is a strong indication that these two paintings were made in both the same workshop and period. Further researches include threads angles [11], [12], since they provide information about the phenomenon known as cusping, i.e., the displacement of the threads from a rectilinear pattern caused by attaching the canvas to a stretcher with tacks before priming the surface. More recently, more complicated weave patterns, like twills, have also been studied [13].

Most of previous studies have used two-dimensional discrete Fourier transforms (2-D DFTs) to develop computer-assisted tools for analyzing canvas weaves [8]-[17]. To summarize, the locations of spectral peaks on the horizontal and vertical axes of the 2-D DFT correspond to the thread densities. In addition, thread tilting from alignment to the axes of the canvas leads to a corresponding rotation of the spectrum. This procedure is performed on a small piece of fabric and repeated until the whole canvas is analyzed, giving maps of local densities and angles for the entire painting.

Other recent research has explored techniques that are more sophisticated, such as autocorrelation and pattern recognition algorithms [18], synchrosqueezed transforms [6] or machine learning models [7]. They try to develop maps of thread patterns that are more detailed. Once the paintings thread maps have been created, we need a method to relate them and help matching paintings. The customary way to do that makes use of two steps. The first one is a comparison of some feature of the weave, such as the mode or mean of the frequencies of the threads. Provided a similarity is found, the second step is to compare some appropriate features, such as the averaged densities [12], average density deviations [13] or the histograms of the cross-correlations [17].

All these methods have proved successful when applied to artworks of Vincent van Gogh, Johannes Vermeer or Diego Velázquez. Nevertheless, they show a poor performance when applied to some other paintings, especially to deteriorated ones. In these ones, fabric is much more irregular and usually exhibits larger twisting due to frames or seams. Also, part of the original canvas may be missing or primer can be thick enough to not let some threads be distinguishable in the X-ray. Therefore, a mere thread count may not provide any consistent result. In addition, thread charts are useful when matching two adjacent canvas but much less appropriate when they come from slices apart in their roll. In particular, if we are dealing with small canvases the probability that they are contiguous pieces of fabric is fewer.

In this paper, we propose a different approach. Rather than finding an extremely exhaustive thread map, we search for a canvas fingerprint that characterizes the whole roll and even the production process. This fingerprint allows to deal with fabrics with more sophisticated patterns.

To characterize the painting, we roughly follow the model proposed in [19], which uses Fourier analysis like the aforementioned papers but from a different point of view. Instead of locating the greatest spectral peak on the horizontal and vertical axes, a wider collection of maxima is found. In this paper we also do that, but the position of these maxima is selected according to a thorough theoretical study that we derive from lattices and apply to weave patterns. In addition, instead of divide the canvas into small pieces and perform the analysis for each one independently, we average the maxima for the entire painting.

Furthermore, we propose to calculate an estimation of the power spectral density (PSD) instead of simply perform a DFT, since it is more appropriate to deal with random signals [20]. More precisely, an averaged periodogram proves a suitable choice. We find that the position of the averaged maxima of the spectral density follows a pattern that is strictly related to the fabric structure. Accordingly, we extract a set of parameters who completely characterize the canvas. This is a fingerprint that keeps quite stable provided the manufacturing technique does not change. This novel method exhibits promising results in paintings by Goya, Velázquez or Rivera.

From the above discussion, the contribution of this paper is threefold. The first one is of theoretical kind. We prove that the Fourier transform of a 2-dimensional periodic structure, such a weave, has a triangular shaped repetitive pattern and we fully characterize it. Up to our knowledge, there is only some rough, not rigorous results about this issue in the literature [19]. The two other contributions of our paper are of practical nature. On the one hand, our PSD method yields some features that help us to characterize the canvases so that an accurate comparison between them is possible, with better results than previous techniques. On the other hand, this PSD analysis will serve also as a counting method, providing reliable results even in situations where the standard DFT-based count fails. Actually, our method has already proven useful in the research at the Museo Nacional del Prado [21].

## 2. Standard DFT method for thread counting

The simplest yet most widely used method for canvas analysis is thread counting. The vertical threads mounted in a loom are called the warp while the horizontal strands are known as the weft. However, the warp and the weft may be associated to either the vertical or the horizontal fibers in the canvas because the artist places the piece of weave on the stretcher in any position.

Thread count data is commonly employed as evidence for dating and attribution of paintings [9]. The horizontal and vertical thread count statistics provides information about the roll from which it was cut. If the statistics of two canvases disagree (even when allowing for rotation), then they most likely have a different origin, whereas if they agree to some tolerance level they could have been cut from the same roll. This helps the art historian to conclude whether two paintings are contemporary with each other [12].

Manual thread counting is a very tedious process. In addition, the strands do not run in precise straight lines and this makes a human measurement difficult. By using signal processing techniques, the hope is to develop algorithms that simplify and enhance manual measurements, allowing a more detailed analysis of paintings.

The standard method to perform an automatic counting lies on the hypothesis that the threads follow an approximately periodic pattern. This pattern may be supposed to follow a sinusoidal model [8], [9] or other shapes like rectangles [13]. In any case, it is assumed that both horizontal and vertical strands have a fundamental frequency of repetition. For that reason, the core of the standard thread counting algorithm is a discrete Fourier transform (DFT).

More precisely, the X-ray image of the canvas is windowed into small swatches (squares of side about 1 cm) and its two-dimensional DFT is computed. To obtain spectral details, the Fourier transform size is much larger than the windowed section. Ignoring the lowest spectral components, which are due to the prevalent grey-colored tones in the painting, several dominant peaks appears in the DFT. However, thanks to the real nature of numbers representing the image, the spectrum is symmetrical. Then, only half of these peaks are distinct. In particular, for a plain-weave the dominant peak in the $x$-axis (vertical threads frequency) corresponds to the vertical threads counting while the main peak in the $y$-axis (horizontal threads frequency) belongs to the horizontal threads counting.

Furthermore, there is no guarantee that the weave will align with the $x$ and $y$ axes, due to fabric imperfections or mounting artefacts such as canvas stretch. This effect will appear in the DFT as a rotation of the spectral peaks by the same angles as the warp and weft. Therefore, we can also measure the angular deviations of the peaks from the axes since they can provide useful information about the canvas.

Then, we compute another swatch shifted in possibly both horizontal and vertical directions. In addition, adjacent swatches are usually overlapped. The process is repeated until the whole canvas is analyzed. Finally, some useful graphs and statistics can be carried out.

As an example of the usefulness of this method, let us compare two paintings by José de Ribera. They are known with certainty to be made at the same time. We can first compare the histograms for both the horizontal and vertical thread countings. They can be seen in Fig. 1(a) and Fig. 1(b) and its resemblance is undoubted, especially in the vertical histogram. In addition, Table 1 presents some important statistical parameters (mode, mean and standard deviation) of the measurement. The resemblance is again quite remarkable, what confirms that the origin of both paintings must be the same roll of fabric.

Despite the usefulness of this method, in some scenarios it may produce misleading results. For instance, let us analyze two paintings attributed to Miguel de Pret. Their histograms are presented in Fig. 2. Besides, we have the statistics of the measurement in Table 1. In this case, there is no resemblance at all neither in the histograms nor in the statistics. However, chemical and artistic analysis along with historical data indicate that the two paintings may be strongly connected.

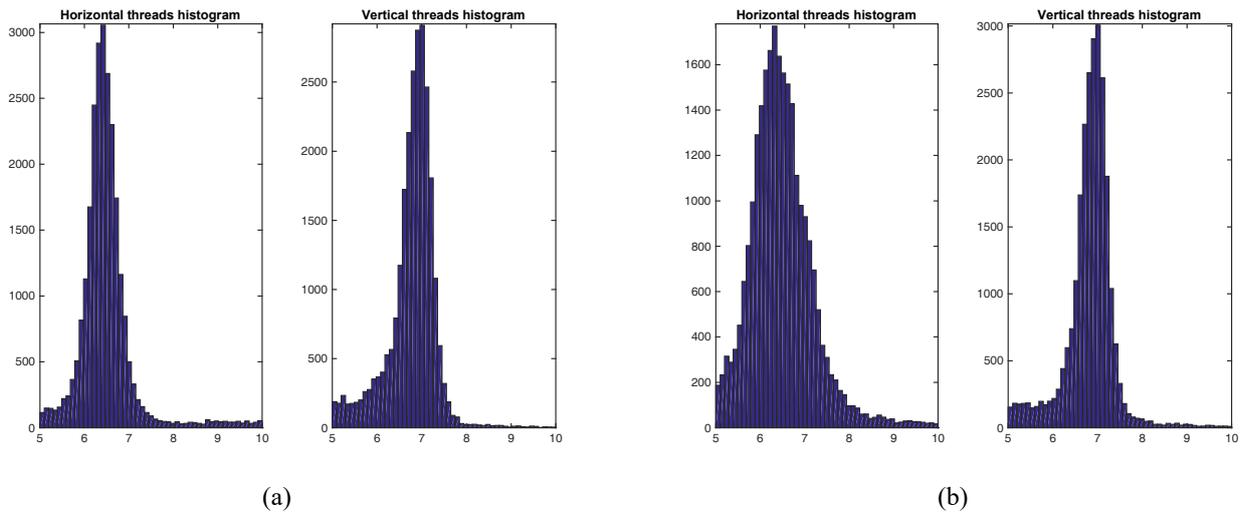

(a)                                                   (b)

Fig. 1: Thread analysis of two paintings by José de Ribera, *Ticio* and *Ixión*, labeled as P01113 and P01114, respectively. The histograms of both the horizontal and vertical thread counting of P01113 are displayed in (a), while (b) contains the histograms of P01114.

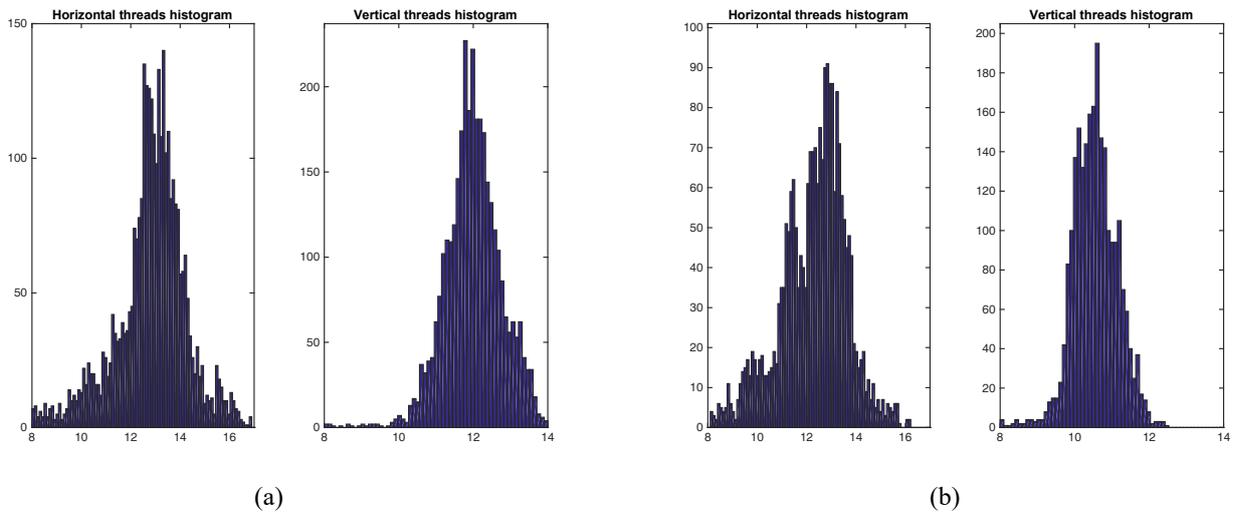

(a)                                                   (b)

Fig. 2: Thread analysis of two paintings attributed to Miguel de Pret: *Two Bunches of Grapes*, labeled as P07905 and *Two Bunches of Grapes with a Fly*, labeled as P07906. Their histograms are displayed in (a) and (b), respectively.

Actually, if we deal with very deteriorate paintings, this standard DFT-based method may fail even in its basic thread counting task. As an example, let us consider an artwork by Goya, labeled as S0A001. Its X-ray is quite difficult to analyze because the prime coat makes several threads look like only one, as seen in Fig. 3(a). Most of the painting is similar to that. The mode of the histogram, shown in Fig. 4, yields 5.15 and 6.81 for vertical and horizontal threads, respectively. On the contrary, Fig. 3(b) shows a very clear area we found within the X-ray, where we can manually check that there are approximately 10 vertical threads and 7 horizontal ones. Henceforth, the thread counting fails.

Table 1: Thread counting statistics for P01113, P01114, P07905 and P07906

|  | Vertical Threads | | | Horizontal Threads | | |
|---|---|---|---|---|---|---|
|  | Mode | Mean | StD | Mode | Mean | StD |
| P01113 | 7.01 | 6.80 | 0.54 | 6.42 | 6.52 | 0.63 |
| P01114 | 7.01 | 6.85 | 0.55 | 6.32 | 6.52 | 0.71 |
| P07905 | 11.79 | 12.01 | 0.76 | 13.33 | 12.81 | 1.45 |
| P07906 | 10.61 | 10.56 | 0.60 | 12.84 | 12.26 | 1.37 |

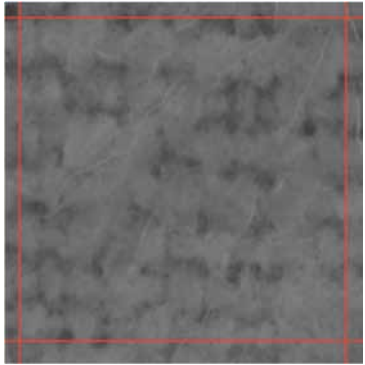
(a)

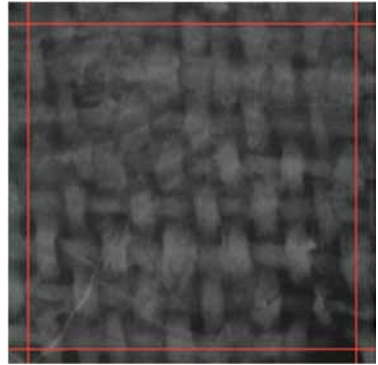
(b)

Fig. 3: Two details at different positions of the X-ray of painting S0A001 by Goya, with a 1 cm grid.

Therefore, our objective will be to find another characterization that is able to count threads and discover similarities between canvases even when the previous methods do not work. To do so, we will search for a fingerprint that portray the whole roll and even the production process.

## 3. Lattices and Fourier transform

### 3.1 Lattices in vector form

Throughout our analysis of the Fourier description of a canvas, we will find out the usefulness of a lattice, i.e., a grid of points that follow a regular pattern. Lattices are customary in fields like crystallography [22] or Fourier optics [23]. Let us first consider a squared lattice $\Lambda_0$, i.e., a set of evenly spaced integer points, as represented in Fig. 5(a).

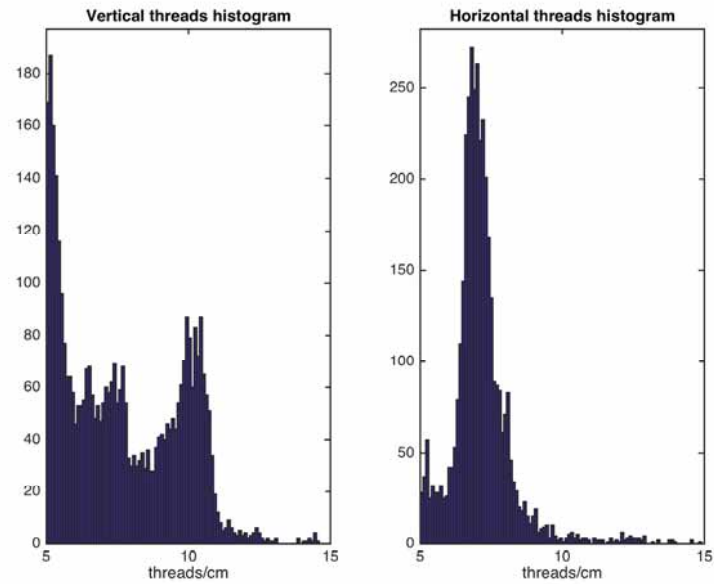

Fig. 4: Histogram for a painting by Goya, labeled as S0A001.

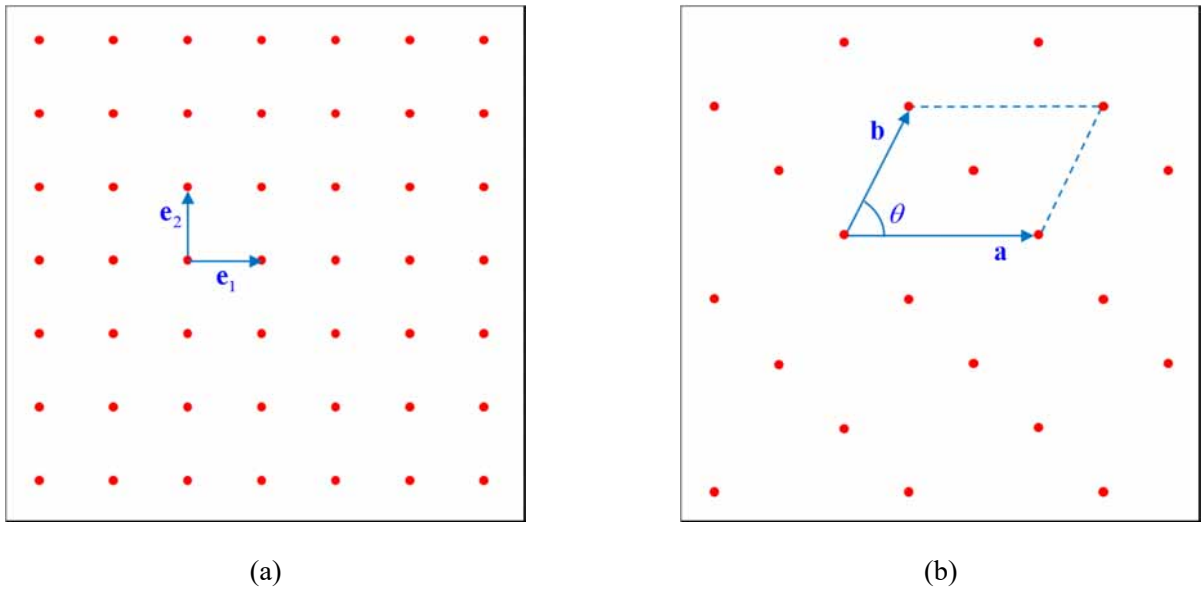

(a)            (b)

Fig. 5: Panel (a) represents a squared integer lattice and the standard vector basis $(\mathbf{e}_1, \mathbf{e}_2)$. Panel (b) shows a general lattice $\Lambda$ and one possible vector basis $(\mathbf{a}, \mathbf{b})$. The angle between $\mathbf{a}$ and $\mathbf{b}$ is $\theta$. The area of the parallelogram is

$$\det(\mathbf{Q}) = \det([\mathbf{a} \mid \mathbf{b}]).$$

Although a lattice can be described in terms of one-dimensional functions, it is much more suitable to use a vector point of view. Lattice $\Lambda_0$ can be represented in terms of vectors by using the standard basis of the plane $\mathbf{R}^2$, i.e., the vectors $\mathbf{e}_1 = [1, 0]^T$ and $\mathbf{e}_2 = [0, 1]^T$, where the superscript T stands for transpose. We can write the set of points in the lattice as

$$\mathbf{p}_0 = n_1 \mathbf{e}_1 + n_2 \mathbf{e}_2, \quad n_1, n_2 = 0, \pm 1, \pm 2, ... \tag{1}$$

Nevertheless, we will have to handle quadrangular lattices with non-squared spaced points and even oblique arrangements, as shown in Fig. 5(b). Let us denote such a lattice by $\Lambda$. For this more general case, take any basis $\mathbf{a} = [a_1, a_2]^T$ and $\mathbf{b} = [b_1, b_2]^T$ of the grid in $\Lambda$ and consider all the vectors that are integer linear combinations of the basis,

$$\mathbf{p} = n_1 \mathbf{a} + n_2 \mathbf{b}, \quad n_1, n_2 = 0, \pm 1, \pm 2, ... \tag{2}$$

Any lattice $\Lambda$ can be obtained from $\Lambda_0$ via an invertible linear transformation, $\mathbf{Q}$. More precisely, we can define a matrix $\mathbf{Q} = [\mathbf{a} \mid \mathbf{b}]$ such that $\mathbf{a} = \mathbf{Q}\mathbf{e}_1$ and $\mathbf{b} = \mathbf{Q}\mathbf{e}_2$, and rewrite the grid in lattice $\Lambda$ from the points in $\Lambda_0$,

$$\mathbf{p} = n_1 \mathbf{Q}\mathbf{e}_1 + n_2 \mathbf{Q}\mathbf{e}_2 = \mathbf{Q}(n_1 \mathbf{e}_1 + n_2 \mathbf{e}_2) = \mathbf{Q}\mathbf{p}_0. \tag{3}$$

Besides, the basis (**a**, **b**) or, in other words, the transformation matrix $\mathbf{Q}$, defines a fundamental parallelogram in $\Lambda$ whose area is $|\det(\mathbf{Q})|$. The basis for $\Lambda$ is not unique but the area remains always the same.

## 3.2 Reciprocal Lattice

Given a lattice $\Lambda$ with points **p**, the reciprocal lattice $\overline{\Lambda}$ consists of all points **q** such that $\mathbf{p} \cdot \mathbf{q}$ is an integer. The linear transformation associated with $\overline{\Lambda}$ is given by $\mathbf{Q}^{-T}$ and one basis for $\overline{\Lambda}$ is $\overline{\mathbf{a}} = \mathbf{Q}^{-T}\mathbf{e}_1$ and $\overline{\mathbf{b}} = \mathbf{Q}^{-T}\mathbf{e}_2$. Then, we can use (3) and (1) to get

$$\mathbf{p} \cdot \mathbf{q} = (\mathbf{Q}\mathbf{p}_0) \cdot (\mathbf{Q}^{-T}\mathbf{p}_0) = \mathbf{p}_0 \cdot (\mathbf{Q}^T\mathbf{Q}^{-T}\mathbf{p}_0) = \mathbf{p}_0 \cdot \mathbf{p}_0 = (n_1 \mathbf{e}_1 + n_2 \mathbf{e}_2) \cdot (n_1' \mathbf{e}_1 + n_2' \mathbf{e}_2) = n_1 n_1' + n_2 n_2', \tag{4}$$

which is an integer. We can follow the same reasoning to proof that

$$\mathbf{a} \cdot \overline{\mathbf{b}} = (\mathbf{Q}\mathbf{e}_1) \cdot (\mathbf{Q}^{-T}\mathbf{e}_2) = \mathbf{e}_1 \cdot (\mathbf{Q}^T\mathbf{Q}^{-T}\mathbf{e}_2) = \mathbf{e}_1 \cdot \mathbf{e}_2 = 0, \tag{5}$$

$$\mathbf{b} \cdot \overline{\mathbf{a}} = (\mathbf{Q}\mathbf{e}_2) \cdot (\mathbf{Q}^{-T}\mathbf{e}_1) = \mathbf{e}_2 \cdot (\mathbf{Q}^T\mathbf{Q}^{-T}\mathbf{e}_1) = \mathbf{e}_2 \cdot \mathbf{e}_1 = 0. \tag{6}$$

This means that the bases of $\Lambda$ and $\overline{\Lambda}$ constitute two pairs of perpendicular vectors, as shown in Fig. 6(a). Moreover, the area of the fundamental parallelogram of $\overline{\Lambda}$ given by (**a**, **b**) will be $|\det(\mathbf{Q}^{-T})| = |\det(\mathbf{Q})^{-1}|$, i.e., the inverse of the area of the parallelogram for the lattice $\Lambda$. In addition,

$$\mathbf{a} \cdot \overline{\mathbf{a}} = (\mathbf{Q}\mathbf{e}_1) \cdot (\mathbf{Q}^{-T}\mathbf{e}_1) = \mathbf{e}_1 \cdot (\mathbf{Q}^T\mathbf{Q}^{-T}\mathbf{e}_1) = \mathbf{e}_1 \cdot \mathbf{e}_1 = 1. \tag{7}$$

Besides, from Fig. 6(a) we have

$$\mathbf{a} \cdot \mathbf{\bar{a}} = |\mathbf{a}||\mathbf{\bar{a}}|\cos\left(\frac{\pi}{2} - \theta\right) = |\mathbf{a}||\mathbf{\bar{a}}|\sin(\theta). \tag{8}$$

Henceforth,

$$|\mathbf{\bar{a}}| = \frac{1}{|\mathbf{a}|\sin(\theta)}, \tag{9}$$

and

$$|\mathbf{\bar{b}}| = \frac{1}{|\mathbf{b}|\sin(\theta)}. \tag{10}$$

In other words, the length of $\mathbf{\bar{a}}$ is the inverse of the distance between $\mathbf{b}$ and its parallel side in the fundamental parallelogram in $\Lambda$, as represented in Fig. 6(b). Analogously, the length of $\mathbf{\bar{b}}$ will be the inverse of the distance between $\mathbf{a}$ and its parallel side.

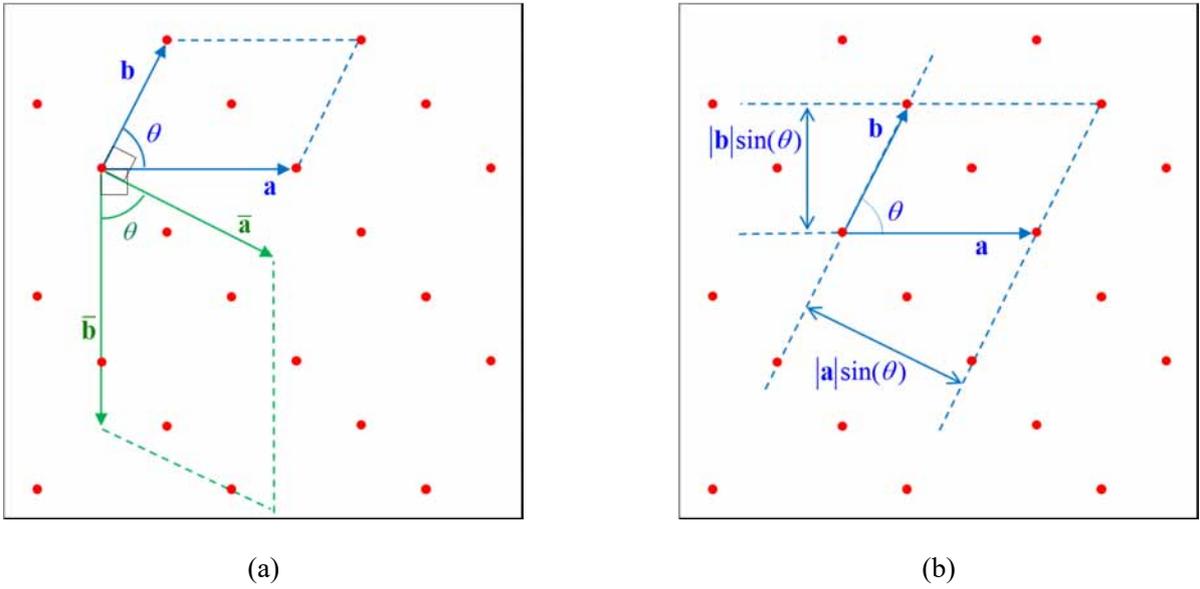

(a)          (b)

Fig. 6: The lattice $\Lambda$, one possible vector basis $(\mathbf{a}, \mathbf{b})$ and the corresponding vector basis $(\mathbf{\bar{a}}, \mathbf{\bar{b}})$ for the reciprocal lattice $\bar{\Lambda}$ are shown in (a). Vector $\mathbf{\bar{a}}$ is perpendicular to $\mathbf{b}$ and vector $\mathbf{\bar{b}}$ is perpendicular to $\mathbf{a}$, where $(\mathbf{a}, \mathbf{b})$ is the vector basis of the original lattice $\Lambda$. Then, the angle $\theta$ between $\mathbf{\bar{a}}$ and $\mathbf{\bar{b}}$ will be the same as the angle between $\mathbf{a}$ and $\mathbf{b}$. The area of the parallelogram in $\bar{\Lambda}$ is the inverse of the area of the one in $\Lambda$. Projections of vectors $\mathbf{a}$ and $\mathbf{b}$ are represented in (b). The inverses of the projection lengths, $|\mathbf{a}|\sin(\theta)$ and $|\mathbf{b}|\sin(\theta)$, are respectively the lengths of vectors $\mathbf{\bar{a}}$ and $\mathbf{\bar{b}}$ in the basis of the reciprocal lattice.

## 3.3 Application to the Fourier transform

Bearing in mind the results above, we can face the Fourier transform (FT) of delta functions. Let us define

$$h(\mathbf{x}) = \sum_{\mathbf{p} \in \Lambda} \delta(\mathbf{x} - \mathbf{p}), \qquad (11)$$

where the variable $\mathbf{x}$ stands for any pair of coordinates in $\mathbf{R}^2$. Then, $h(\mathbf{x})$ represents a set of impulses situated on the points of a lattice $\Lambda$. The reciprocal set of impulses corresponding to the reciprocal lattice $\bar{\Lambda}$ is

$$\bar{h}(\mathbf{x}) = \sum_{\mathbf{q} \in \bar{\Lambda}} \delta(\mathbf{x} - \mathbf{q}). \qquad (12)$$

We bring here a result from [24], which shows that the FT of $h(\mathbf{x})$ in (11) is

$$H(\mathbf{f}) = \mathcal{F}\{h(\mathbf{x})\} = \left|\det(\mathbf{Q}^{-T})\right| \cdot \bar{h}(\mathbf{f}) = \frac{1}{\left|\det(\mathbf{Q})\right|} \cdot \sum_{\mathbf{q} \in \bar{\Lambda}} \delta(\mathbf{f} - \mathbf{q}), \qquad (13)$$

where $\mathbf{f} = (f_1, f_2)$ is a two-dimensional frequency variable. Put in other words, the FT of a set of impulses corresponding to the lattice $\Lambda$ is another set of impulses located at the points of the reciprocal lattice $\bar{\Lambda}$.

## 3.4 Relation between lattices

From the previous subsections, we know the relation between the vector basis of a lattice and its reciprocal one. However, we need a more general result to analyze more complex patterns, like the ones we find in fabrics. Let us derive this novel result.

If we express $\mathbf{Q}$ in terms of its coefficients,

$$\mathbf{Q} = [\mathbf{a} \mid \mathbf{b}] = \begin{bmatrix} a_1 & b_1 \\ a_2 & b_2 \end{bmatrix}, \qquad (14)$$

$\mathbf{Q}^{-T}$ yields

$$\mathbf{Q}^{-T} = \frac{1}{a_1 b_2 - a_2 b_1} \begin{bmatrix} b_2 & -a_2 \\ -b_1 & a_1 \end{bmatrix} = \frac{1}{\det(\mathbf{Q})} \cdot \begin{bmatrix} b_2 & -a_2 \\ -b_1 & a_1 \end{bmatrix}. \qquad (15)$$

In addition, $\mathbf{a}' = [-b_1, -b_2]^T$ and $\mathbf{b}' = [a_1, a_2]^T$ is obviously another valid basis for the lattice $\Lambda$. Let us denote $\mathbf{Q}' = [\mathbf{a}' \mid \mathbf{b}']$ and apply a matrix rotation $\mathbf{R}$ of 90 degrees to this new basis. Then,

$$\mathbf{R}\mathbf{Q}' = \begin{bmatrix} 0 & -1 \\ 1 & 0 \end{bmatrix} \begin{bmatrix} -b_1 & a_1 \\ -b_2 & a_2 \end{bmatrix} = \begin{bmatrix} b_2 & -a_2 \\ -b_1 & a_1 \end{bmatrix}, \qquad (16)$$

which, combined with (15), yields

$$\mathbf{Q}^{-T} = \frac{1}{\det(\mathbf{Q})} \cdot \mathbf{R}\mathbf{Q}'. \qquad (17)$$

Otherwise stated, the reciprocal lattice is merely a rotation of 90 degrees of the original one. Moreover, since

$$\det(\mathbf{Q}^{-T}) = \frac{1}{[\det(\mathbf{Q})]^2} \cdot \begin{vmatrix} b_2 & -a_2 \\ -b_1 & a_1 \end{vmatrix} = \frac{1}{\det(\mathbf{Q})}, \qquad (18)$$

the reciprocal lattice is scaled in a way that the area of its fundamental parallelogram is the inverse of the initial area. This means, among other things, that angles and proportions are the same in both the original and reciprocal lattices and therefore, by virtue of Section 3.3, they are the same in both space and frequency domains. This result, which imposes a tight geometrical relationship between both domains, will play a fundamental role in the subsequent discussion.

## 4. Spectral Analysis: Theory and Practice

### 4.1 Model of the weave pattern

In this section we propose a unified signal processing model for the fabric, comparing both space and frequency domains. A similar model was outlined in [19], but no theoretical proof was provided. In [13], results for some types of cloth are derived by computing the FT for every scenario. Our analysis will be more general, easily allowing the derivation of every result in [13] from our single model.

In an ideal case, a woven fabric has a periodic structure produced by the interleaved pattern of horizontal and vertical threads. This can be described as a convolution of a basic thread shape, $b(\mathbf{x})$, and a field of impulses $h(\mathbf{x})$ that creates the repetition pattern, as shown in Fig. 7. Put in other words,

$$i(\mathbf{x}) = b(\mathbf{x}) * h(\mathbf{x}) = \sum_{\mathbf{p} \in \Lambda} b(\mathbf{x} - \mathbf{p}), \qquad (19)$$

where $i(\mathbf{x})$ is the image intensity. The basic shape $b(\mathbf{x})$ gives information about the threads characteristics and is the more elementary unit in the weave. This basic shape will be repeated at evenly spaced points by means of the convolution with $h(\mathbf{x})$ [19].

Although we have different choices for $b(\mathbf{x})$ and for the vector basis, for the forthcoming analysis becoming simpler we take $b(\mathbf{x})$ as a horizontal rectangle and $\mathbf{a}$ as a horizontal vector, as in Fig. 7(b) and Fig. 7(c). This will be always possible due to the vertical and horizontal arrangement of the fibers. In addition, the two vectors $\mathbf{a}$ and $\mathbf{b}$ will be assumed to span an angle less or equal than 90 degrees. Moreover, if we take $d_v$ as the distance between two consecutive vertical threads and $d_h$ as the distance between two consecutive horizontal threads, $\mathbf{a}$ and $\mathbf{b}$ will be of the form

$$\mathbf{a} = [md_v, 0]^T, \qquad (20)$$

$$\mathbf{b} = [nd_v, pd_h]^T, \qquad (21)$$

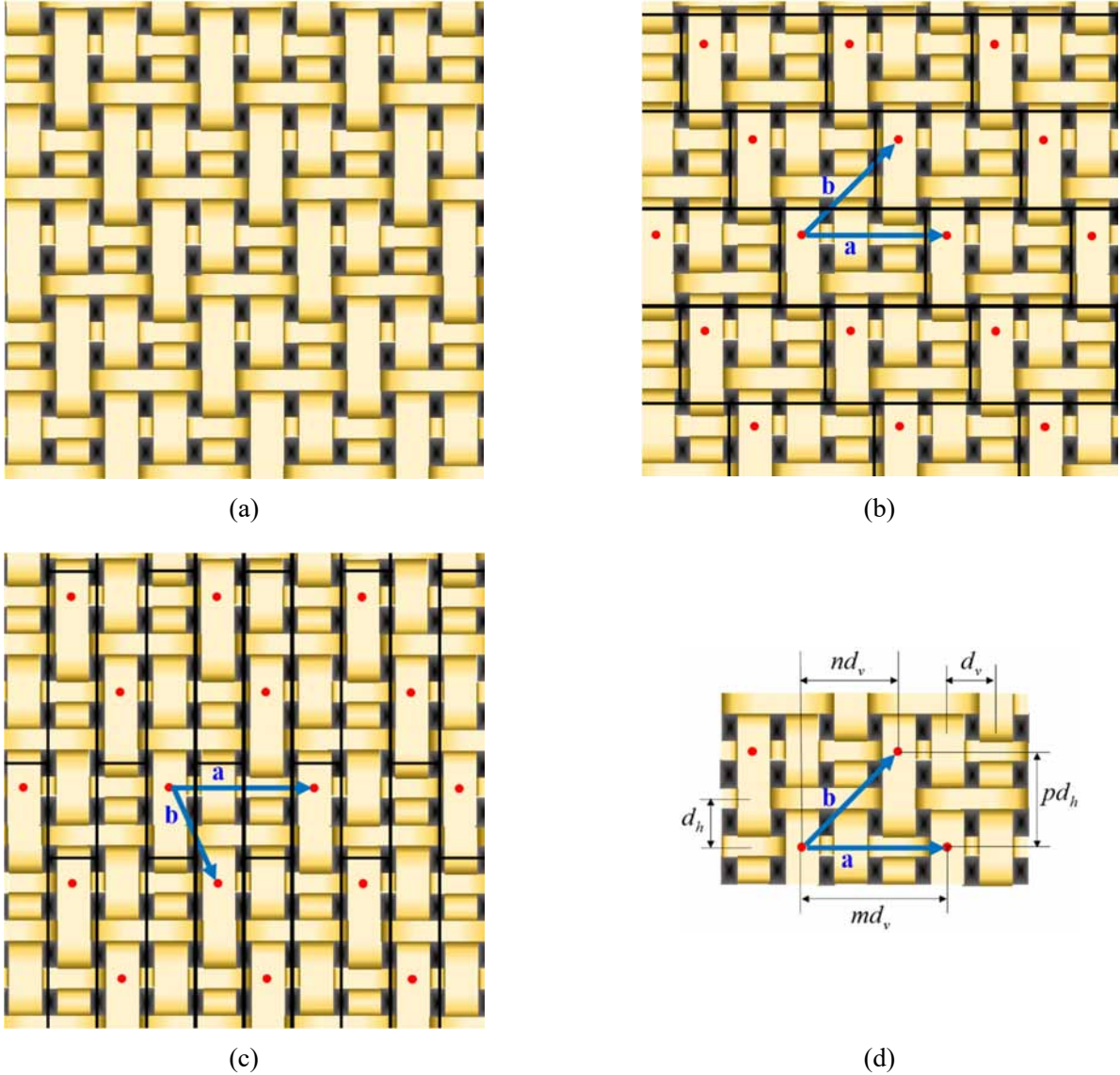

Fig. 7: Model of the repetition pattern of the weave image $i(\mathbf{x})$ shown in (a). Two different choices for the basic shape $b(\mathbf{x})$ and vector basis ($\mathbf{a}$, $\mathbf{b}$) are presented in (b) and (c). The dots are the points of the lattice. Its corresponding grid of delta functions is $h(\mathbf{x})$. A rectangle representing the basic shape $b(\mathbf{x})$ is associated to each dot. The rectangles and the vector $\mathbf{a}$ are horizontal, in both (b) and (c). Panel (d) displays a swatch explaining the parameters that describe the weave in (b), where $m = 3$, $n = 2$, $p = 2$.

where $m$, $n$ and $p$ are integers that tell us the number of strands involved in the basic shape. In particular, the basic shape, $b(\mathbf{x})$, will contain $p$ horizontal threads and $m$ vertical threads. Henceforth, in order to describe completely the weave, we need to know $m$, $n$, $p$, $d_h$ and $d_v$. For instance, plain-weaves have $m = 2$ and $n = 1$, and a basic twill fabric,

$m > 2$ and $n = 1$ or $n = m - 1$, as shown in Fig. 8.

Provided this condition, we can build right triangles like those in Fig. 9 (a) and Fig. 9 (b). In particular, the number of segments of length $|\mathbf{b}|$ in the hypotenuse is $m$ whereas the number of segments of length $|\mathbf{a}|$ in the horizontal leg is $n$. Besides, the number of segments in the opposite leg will be always one. Fig. 9(c) and Fig. 9(d) shows the most important characteristics of these triangles.

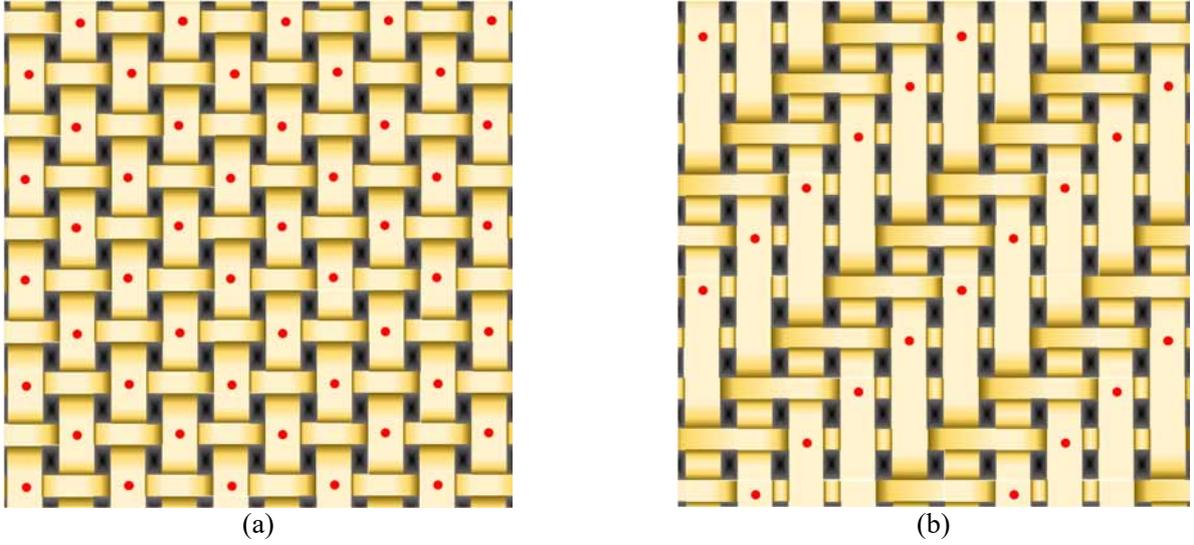

(a)          (b)

Fig. 8: Panel (a) shows a plain-weave. Its pattern parameters are $m = 2, n = 1, p = 1$. Panel (b) is an example of a simple twill. Its fabric parameters are $m = 5, n = 1, p = 1$.

### 4.2 Spectral analysis of the weave

Now, we recall the convolution theorem and from (19) we calculate the $N_{DFT}$-point Fourier transform of the image intensity as

$$I(\mathbf{f}) = B(\mathbf{f})H(\mathbf{f}), \tag{22}$$

where $B(\mathbf{f})$ is the FT of the basic shape. Using (13), $I(\mathbf{f})$ can be expressed as

$$I(\mathbf{f}) = \frac{1}{|\det(\mathbf{Q})|} B(\mathbf{f}) \cdot \sum_{\mathbf{q} \in \Lambda} \delta(\mathbf{f} - \mathbf{q}) = \frac{1}{|\det(\mathbf{Q})|} \cdot \sum_{\mathbf{q} \in \Lambda} B(\mathbf{q})\delta(\mathbf{f} - \mathbf{q}). \tag{23}$$

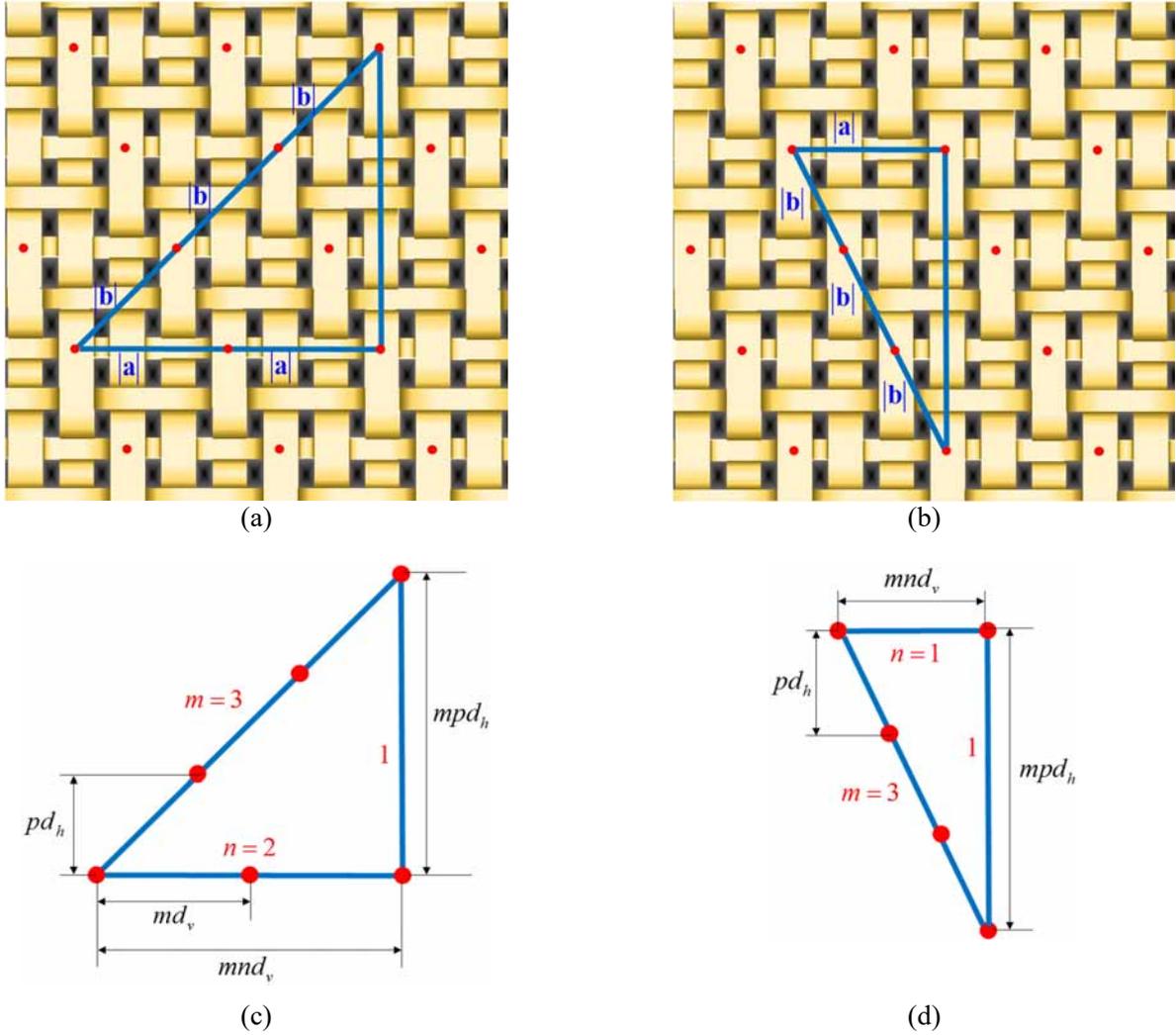

Fig. 9: Two fundamental right triangles in the pattern of a weave image $i(\mathbf{x})$ are shown in (a) and (b), corresponding to Fig. 7(b) and Fig. 7(c), respectively. Panels (c) and (d) display the number of segments spanned in each side and the main dimensions of the triangles.

This means that the spectrum of the image intensity has components only in the frequencies determined by the grid of the reciprocal lattice. In addition, by virtue of (23), each point in the frequency-domain grid is scaled by the value of the spectrum of the basic shape at this particular frequency, as shown in Fig. 10. For its part, from our analysis in Section 3.4 we know that angles and proportions are the same in both space and frequency domains. This implies that, in the frequency domain, there must be a right triangle similar to the one already defined in the space-domain. As a matter of fact, we can observe that the triangle in Fig. 10(e) is a scaled 90º-degree rotation of the one in Fig. 9 (a), i.e., the triangles of both the weave and its Fourier transform are mathematically similar. We can use this result to infer the main characteristics of the weave pattern.

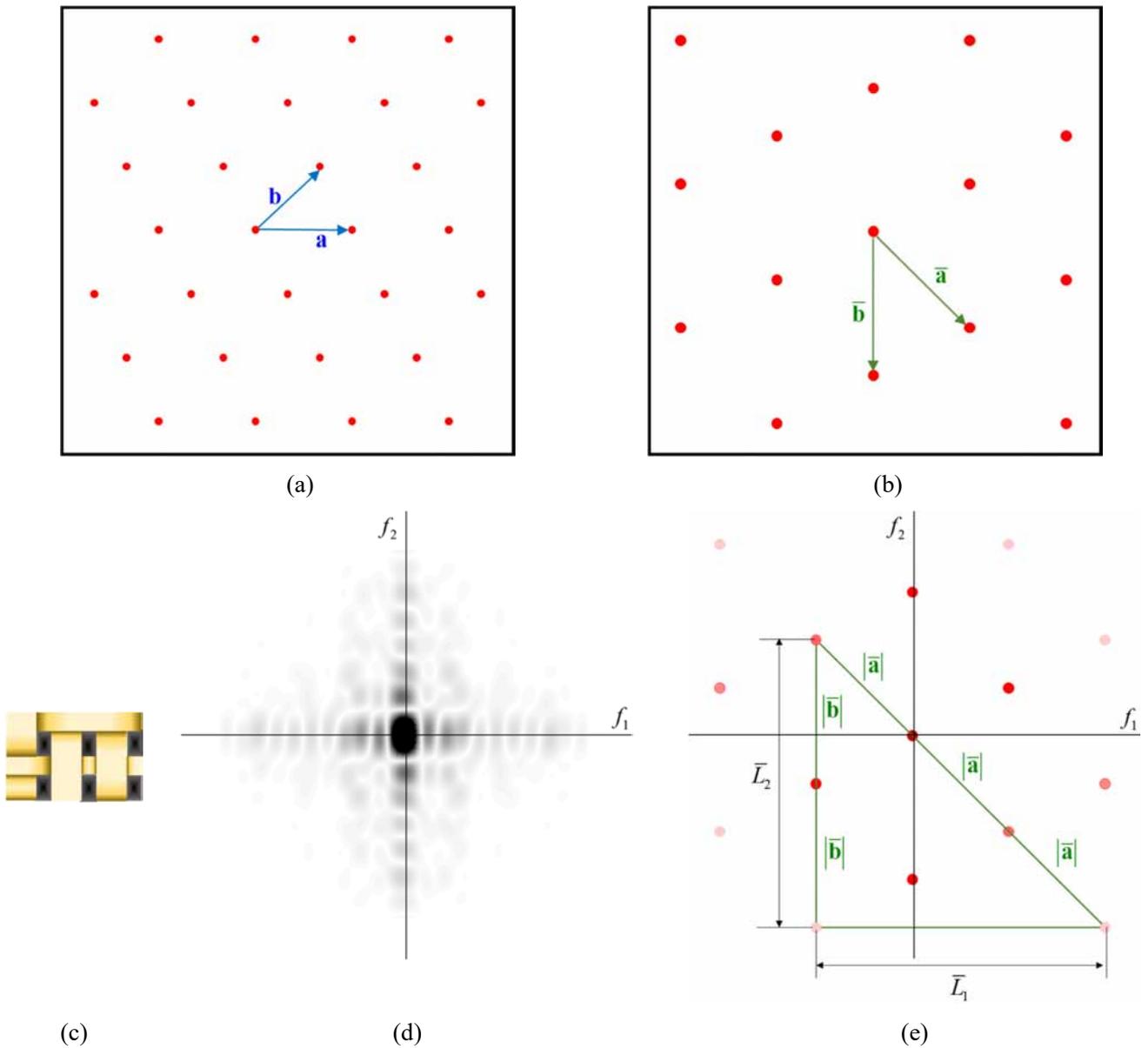

Fig. 10: The lattice from the weave pattern in Fig. 7 is represented in (a). A possible vector basis is also displayed. The reciprocal lattice with the corresponding vector basis, i.e., the Fourier transform of (a), is shown in (b). Panel (c) is a swatch from the canvas showing the basic shape of Fig. 7(b) and panel (d) is the 2-D spectral magnitude of that basic shape. The darker the image is, the higher the magnitude will be. The 2-D spectral magnitude of the weave, i.e., the product of figures in (b) and (d), is represented in (e), not at the same scale than (d). The fundamental right triangle for the reciprocal lattice is also shown.

First, we look for a right triangle in the grid formed by the spectral peaks. Therefore, from the previous discussion, we find the values of *m* and *n* by counting the number of segments in the hypotenuse and the vertical leg of this frequency-domain triangle, respectively.

Second, we can easily measure the sides lengths $\bar{L}_1$ and $\bar{L}_2$ (see Fig. 10) and derive the values for the thread frequencies, $f_h$ and $f_v$, as follows. From (9), (20) and the triangles in Fig. 9(c) and Fig. 10(e), we could write

$$|\mathbf{a}|\sin(\theta) = \frac{\bar{L}_1}{m} = \frac{1}{|\mathbf{a}|} = \frac{1}{md_v}. \tag{24}$$

Hence, we can calculate the vertical thread frequency as

$$f_v = \frac{1}{d_v} = \bar{L}_1. \tag{25}$$

On the other hand, from (10), (21) and the triangles in Fig. 9(c) and Fig. 10(e),

$$|\mathbf{b}| = \frac{\bar{L}_2}{n} = \frac{1}{|\mathbf{b}|\sin(\theta)} = \frac{1}{pd_h}. \tag{26}$$

Therefore, the horizontal thread frequency will be

$$f_h = \frac{1}{d_h} = \frac{p\bar{L}_2}{n}. \tag{27}$$

We need the value of *p* to estimate $f_h$. However, we are aware that $p=1$ for most usual fabrics (e.g., plain-weaves and simple twills) and therefore $f_h$ would be uniquely determined.

### 4.3    Power spectral density: the averaged periodogram.

The previous results provide the theory that predicts how the FT of the fabric would be. However, in practice it is usually far from the theoretical result developed above. We estimate it from the X-ray images of the canvases but they do not present a uniform pattern along the warp and weft, i.e., parameters defining the weave vary, especially for old fabrics. In addition, part of the painting is superimposed and there are artifacts such as nails, patches or missing parts. Finally, the theoretical studies assume infinite length sources, which obviously is not true. Therefore, the estimation of the spectral analysis of the weave is not straightforward. Fig. 11 shows an example where we cannot clearly observe the theoretical pattern in the DFT of a window, including maxima located in a grid, i.e., the weave pattern.

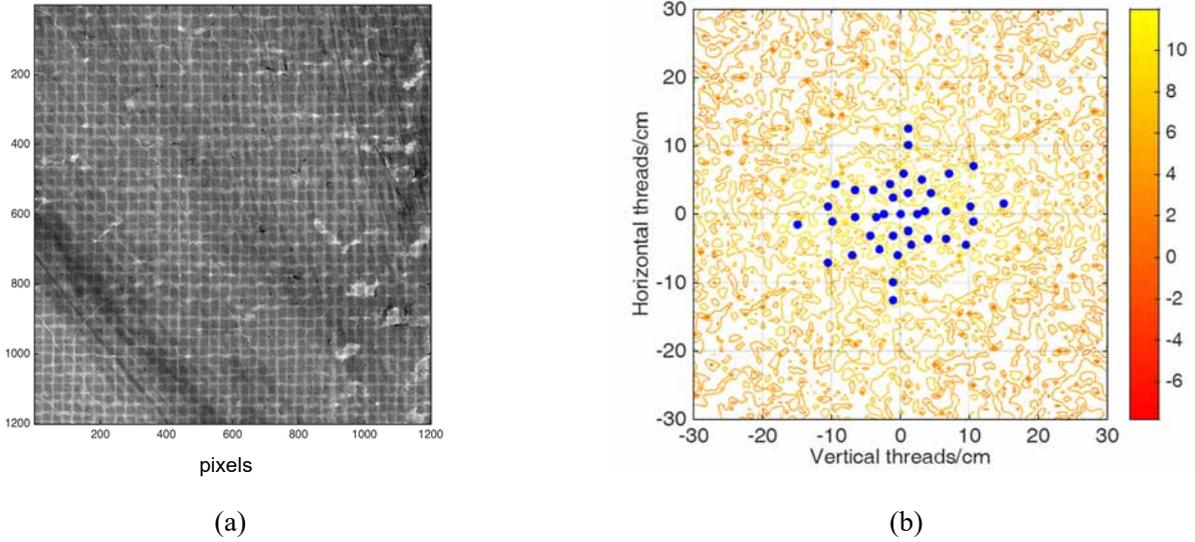

(a)            (b)

Fig. 11: Panel (a) show an X-ray of a swatch in the painting P01134, axes in pixels, and panel (b) is a contour plot of the 400-point 2D-DFT of a 400×400 pixels window with local maxima in dots.

Therefore, the variability of the handcrafted weave process, the placing of the canvas on the stretcher or even the painting itself may alter substantially the thread counting comparison between two fabrics even if they actually came from the same roll. In addition, if the sizes of two canvases are not big enough to facilitate that they come from the same part of the warp, the matching is not possible by spatial correlation of the countings. For all the above reasons, the standard DFT-based methods of previous papers may fail, as discussed in Section 2, because they amount to focusing only on the two main maxima of the DFT to perform the counting.

In this paper, we propose a novel approach in which, rather than finding the maxima of the DFT of a window, we estimate the whole power spectral density (PSD) of the fabric. Actually, we consider the canvas as a statistically nonstationary signal, since its properties vary with the actual swatch. Consequently, we compute the power spectral density of the image rather than the magnitude of the DFT. It is well known that this is the correct approach when nonstationary signals are involved [20]. More precisely, we resort to the averaged periodogram, i.e., we average the squared module of the DFT for several swatches through the X-ray of the canvas. The PSD provides a fingerprint of the canvas as a whole, i.e., it is an additional approach where we focus on a general description of the painting. In this way, we will be able to extract some important features, which will help us to compare canvases where the thread count offers inconclusive results. In particular, we will be able to infer the weave pattern (as explained in Section 4.1) from this averaged spectrum. This is a characteristic that cannot be deduced from previous techniques.

Taking for now the periodogram $S(\mathbf{f})$ as a first PSD estimate, we have that (23) turns into

$$S(\mathbf{f}) = |I(\mathbf{f})|^2 = \frac{1}{|\det(\mathbf{Q})|^2} \cdot \sum_{\mathbf{q} \in \Lambda} |B(\mathbf{q})|^2 \delta(\mathbf{f} - \mathbf{q}). \qquad (28)$$

Therefore, the theoretical PSD of the image intensity has its components only in the frequencies corresponding to the grid of the reciprocal lattice, as the magnitude of the DFT had. For this reason, the analysis of Section 4.2 is completely valid now, too. However, the periodogram is not a consistent estimate, i.e., its variance does not asymptotically approach zero with increasing signal length. Hence, we would rather use periodogram averaging. The averaging is used to decrease the variance of the spectral estimation [20].

More precisely, the image is divided into segments of $N \times N$ samples, with a window $w(n_1, n_2)$ applied to each. The displacement between two consecutive segments is $D$, i.e., we form the $r$-th segment $i_r(n_1, n_2)$ as

$$i_r(n_1, n_2) = i(n_1 + rD, n_2 + rD) w(n_1, n_2), \ 0 \leq n_1, n_2 \leq N-1. \qquad (29)$$

If $D < N$ the segments overlap whereas they are contiguous if $D = N$. Hence, the overlapping ratio $s$ will be

$$s = \frac{N-D}{D}. \qquad (30)$$

The total number of segments $K$ depends on the relationship among $D$, $N$ and the size of the image.

Then, the discrete periodogram of the $r$-th segment is

$$S_r(k_1, k_2) = \frac{1}{NU} |I_r(k_1, k_2)|^2, \qquad (31)$$

where $I_r(k_1, k_2)$ is the $N_{DFT}$-point 2D-DFT of that segment and $U$ is a normalizing constant. Nevertheless, we can always choose $U = 1$ and rest unworried, since we are not interested in the exact value of $S_r(k_1, k_2)$ but in the position of its maxima.

Finally, we compute the averaged periodogram as

$$\tilde{S}(k_1, k_2) = \frac{1}{K} \sum_{r=1}^{K} S_r(k_1, k_2). \qquad (32)$$

The main advantage of this averaging is that the variance of this PSD estimator is inversely proportional to the number of segments. Thus, the variance tends to zero when $K$ increases. In practice, averaging helps to reduce the contribution of deteriorate parts in the canvas, deformed areas or the painting itself. This yields a quite robust estimator. In addition, this method can be readily computed in parallel with the regular thread counting at virtually no additional cost, because both of them share the main core computation, i.e., the DFT of the windowed segments.

## 4.4 Extraction of features

Once we have performed the spectrum analysis of the weave through the PSD, we focus on its characterization, which is not straightforward. Theoretically, the PSD should be a weave pattern with just some non-null points. In practice, we have a piece of fabric of finite length. The windowing used in the PSD computation is introduced to deal with this problem. As a result, the PSD is no more a set of non-null points but a full surface with maxima in the weave pattern. Besides, the averaging reduces the variance of the estimation but systematic imperfections, possibly caused by the loom design or the quality of the used threads, will be still present in the PSD.

Through experimentation, we found that the following four PSD features help to characterize the type of fabric:

1. First, we propose to observe the *shape* of the PSD at distant points of the center, whether it is diamond or cross-like. In Fig. 12, we include two examples of PSD computed with a 2048-point DFT of $N = 400$ pixels patches shifted $D = 100$ points with a Blackman-Harris window. In Fig. 12(a) we have a diamond shape, while in Fig. 12(b) a cross one. The shapes have been highlighted in thick dashed lines. The first one fits better with the theoretical result, as we have points and concentric contour lines around them where maxima points are predicted to be. On the contrary, the contour lines are elongated in Fig. 12(b). We conjecture that this elongation is due to imperfections of the cloth.

2. Then, as a second feature, we pay attention to the horizontal and vertical contour curves connecting the maximum points (marked with dots) in the *diagonal* to the vertical and horizontal axes. In Fig. 12(a) we include a triple-cross shape, in thin dashed line along the axes plus two horizontal lines. In Fig. 12(b) we have a simple cross, in thin dashed line along the axes.

3. A third characteristic to evaluate for is the shape of the contour curves of the PSD near the *center*. In Fig. 13(a) we find a "C" shape, while In Fig. 13(b) there is an "O" shape, meaning that a minimum is located between the center and the first maximum in the diagonal. For its part, in Fig. 13(c) we have a "plain" or "\" type shape, where the PSD is equally decreasing from the center in any direction.

4. Finally, we propose to use the values along the *axes*. More precisely, we may find large values along the whole vertical or horizontal axes or just at some points of them. For instance, in Fig. 12(b) the PSD has a strong value along the vertical axis.

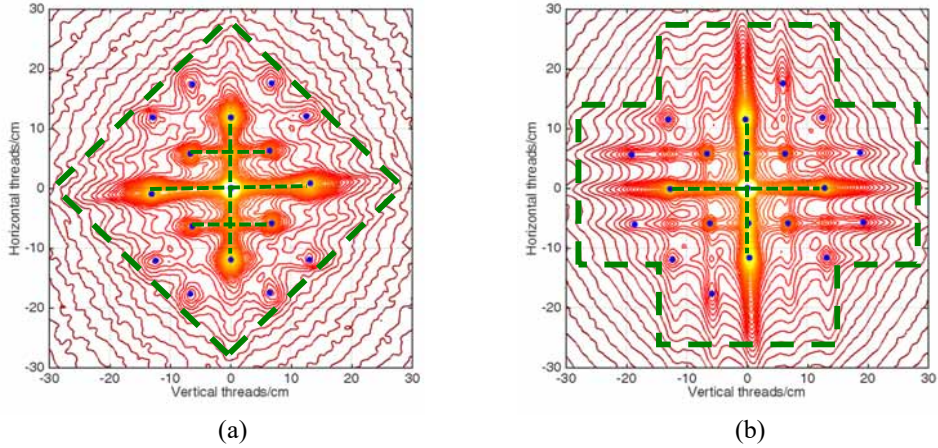

Fig. 12: Shape of the PSD in the edges as a feature of the fabric, (a) diamond shape and (b) cross shape. Also, connection between the maximum points in the diagonal and the *x* and *y* axes is shown.

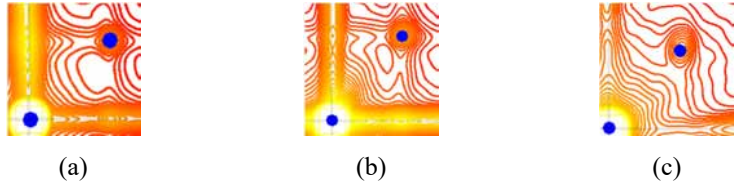

(a)　　　　　　　　　(b)　　　　　　　　　(c)

Fig. 13: Examples of shapes of the PSD near the center as a fabric feature, (a) ″C″ type, (b) ″O″ type and (b) ″\″type.

## 5. Results

### 5.1 Examples of PSD and variance reduction

We first propose to illustrate the technical description of the theoretical Fourier transform of fabrics in Section 4.2 by using the PSD. We include here two examples. First one is a plain-weave by Rivera. A detail of its X-ray is included in Fig. 14(a), where 200 pixels correspond to one cm. Its PSD estimation is shown in Fig. 14(b). Second one is a simple twill by Claudio de Lorena. Its spectral analysis is represented in Fig. 14(d), with the X-ray detail to the left. We mark with dots the local maxima found where the theoretical results predict them to be and plot the theoretical triangles describing the fabric. It is interesting to remark the clean and neat results of the PSD despite distortions and the painting itself. At this point, it is helpful to compare the result of the PSD to the DFT for a window. In Fig. 11(a) we included the same patch as that in Fig. 14(a) and the DFT of a 400×400 pixels window was shown in Fig. 11(b). Compared to Fig. 14(b), it exhibits a quite different result from the theoretical DFT we should get for this fabric.

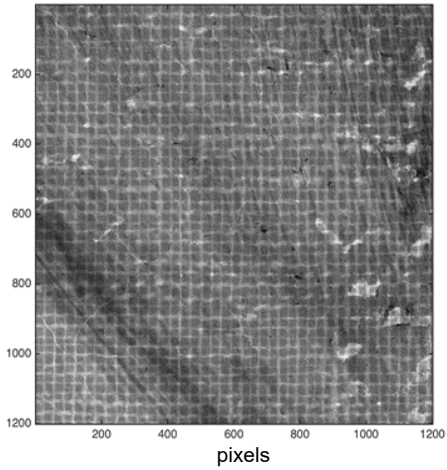

(a)

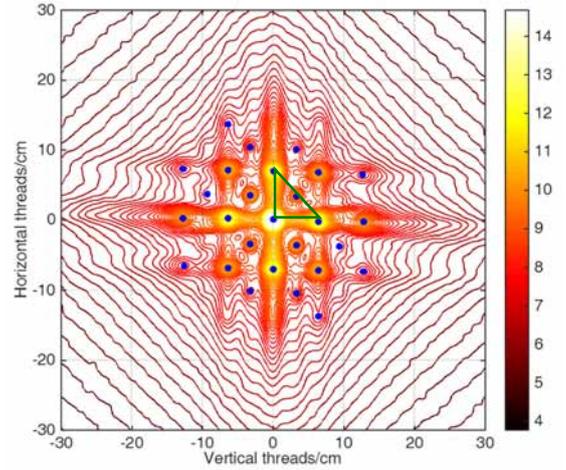

(b)

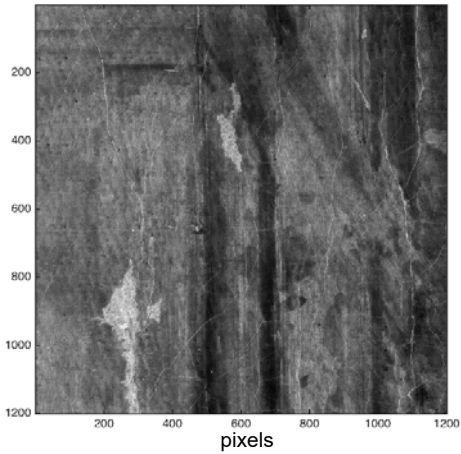

(c)

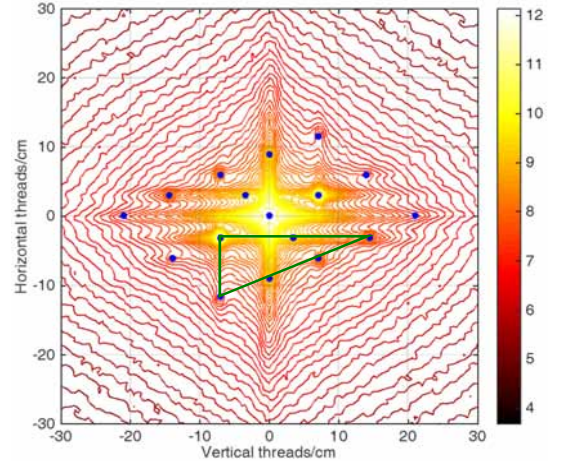

(d)

Fig. 14: Detail of the X-ray of the plain-weave P01113 by Rivera in (a) and the contour plot of its PSD in (b), with $N = 400$, $N_{\text{DFT}} = 2048$ and $D = 100$. Panel (c) represents a portion of the radiography of the twill P02254 by Claudio de Lorena and the contour plot of its PSD is in (d), with $N = 400$, $N_{\text{DFT}} = 400$ and $D = 50$ (d). Triangle patterns and a set of local maxima (dots) are included.

## 5.2 Thread Counting with PSD

The PSD can be used for thread counting by simply measuring the dimensions of the spectral triangle described in the theoretical results of our study in Section 4.2. For instance, the triangle for a plain-weave has its legs parallel to the axes, with a vertex in the origin and two segments in the hypotenuse. Then, the distance to the horizontal and vertical two other vertices provide the thread counting for the vertical and horizontal directions, respectively. For a plain-weave it amounts, most of the times, to find the maxima in the vertical and horizontal axes. In Table 2, we include the results of the thread counting by using histograms of a standard DFT analysis for several paintings. The first six ones are from the 17th century and of plain-weave cloth type. We include a pair of canvases, P07905 and P07906, that were recently studied at the Museo Nacional del Prado to establish their authorship; then, one by Rubens, P01627; one by Teniers, P01819; and two by Velázquez, P01182 and P03253. The last painting in Table 2 is S0A001, by Goya.

In Table 3, we include the result for the thread counting for the same paintings by using PSD. Although similar values are obtained in general, there are some cases where we can observe deviations. The second largest difference can be observed for P01819, which presented the largest StD. This fabric is hard to detect as the painting on top of it has similar frequency components.

Table 2: Thread counting with the DFT method for several paintings of the 17th and 18th centuries.

|  | Vertical Threads | | | Horizontal Threads | | |
| --- | --- | --- | --- | --- | --- | --- |
|  | Mode | Mean | StD | Mode | Mean | StD |
| P07905 | 13.33 | 12.81 | 1.45 | 11.79 | 12.01 | 0.76 |
| P07906 | 12.84 | 12.26 | 1.37 | 10.56 | 10.56 | 0.60 |
| P01627 | 12.77 | 12.53 | 0.97 | 11.70 | 11.60 | 0.82 |
| P01819 | 10.59 | 10.72 | 1.10 | 13.43 | 12.25 | 1.96 |
| P01182 | 11.00 | 10.69 | 0.92 | 10.61 | 10.31 | 0.66 |
| P03253 | 10.61 | 10.65 | 0.89 | 11.79 | 11.61 | 1.27 |
| S0A001 | 5.15 | 7.70 | 2.03 | 6.81 | 7.16 | 1.08 |

Table 3: Thread counting with the PSD method for several paintings of the 17th and 18th century.

|  | Vertical | Horizontal |
| --- | --- | --- |
| P07905 | 13.21 | 11.81 |
| P07906 | 12.69 | 10.61 |
| P01627 | 12.79 | 11.63 |
| P01819 | 10.84 | 13.48 |
| P01182 | 10.93 | 10.36 |
| P03253 | 10.64 | 11.82 |
| S0A001 | 9.96 | 6.95 |

We now focus on painting S0A001, which presents by far the largest deviation. This is a quite difficult radiography to analyze because the prime coat makes several threads look like just one, as seen in Fig. 3. Actually, most of the $N \times N$ swatches are similar to the one in Fig. 3(a). We also recall that for Fig. 3(b) the thread counting is approximately 10 for vertical threads and 7 for the horizontal ones, whereas from the histogram in Fig. 4 the mode is 5.15 and 6.81, respectively. As a matter of fact, the histogram of the vertical counting is quite confusing.

In Fig. 15 we include the PSD computed with a 2048-point DFT and an $N = 400$ pixels Blackman-Harris window. We depict as dots all the local maxima found. Since it is a plain-weave type, we look for the triangle predicted by the theoretical results. Their vertices provide the thread counting as the distance of them to the origin. The triangle has been highlighted in Fig. 15, giving a thread counting of 9.96 and 6.95 for vertical and horizontal threads, respectively. With this triangle, where the maximum in the hypotenuse is taken into account, we discard local maxima in the PSD to provide a correct thread counting.

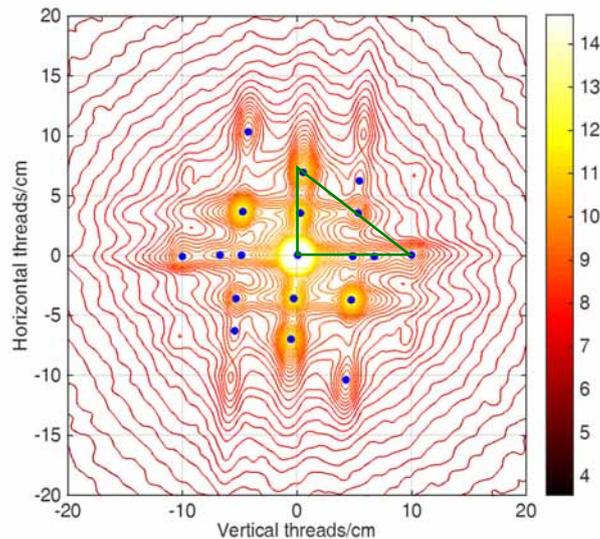

Fig. 15: PSD for S0A001.

## 5.3  Features in PSD

Finally, we propose to use the features in Section 4.4 to classify the fabrics of the first six paintings in Table 2. From it, we can conclude that all countings are in the range 10.61-13.43 threads/cm. First, if we focus on the vertical or horizontal mode value with minimum StD for each painting, and we assume a difference of one thread per cm as a rule to decide for similarity, we found that we can gather P07906, P01819, P01182 and P03253 in one group for their quite close threads counting, (and P07905, P01627 in another). Similarly, studying the counting with larger StD, paintings P07905, P7906, P01627 and P01819 could be related. Therefore, we could conclude that all belong to the same fabric, which does not make sense since some of them were painted by different authors.

We have included the PSD for these paintings in Fig. 16. First, we check the *shape* of the PSD at distant points of the center. We find that it is diamond shaped in Fig. 16(a), Fig. 16(b), Fig. 16(e) and Fig. 16(f) for P07905, P07906, P01182 and P03253, respectively. On the contrary, it is cross-like in Fig. 16(c) and Fig. 16(d), for P01627 and P01819, correspondingly.

Then, we pay attention to the second feature, the horizontal and vertical contour curves connecting the maximum points (marked with dots) in the *diagonal* to the vertical and horizontal axes. It can be observed that Fig. 16(a) and Fig. 16(b) have a large number of them in the horizontal, generating a particular double-cross shape. A third characteristic to evaluate for is the shape of the contour curves of the PSD near the center. While in Fig. 16(a), Fig. 16(b) and Fig. 16(c) we find a "C" shape, in Fig. 16(d) we have "O" shape, and in Fig. 16(e) and Fig. 16(f) we have a "plain" or "\" type shape. Finally, we check for the values along the vertical and horizontal *axes*, where we find that in Fig. 16(c) and Fig. 16(d) the PSD exhibits a high value along the vertical and horizontal axis, respectively, while in Fig. 16(e) and Fig. 16(f) the values along the axes are not as high as at the maxima locations.

The result of the analysis of these features for the paintings in Fig. 16 is summarized in Table 4. From this table we conclude that P07905 and P07906 share the same diamond shape in the edges as P01182 and P03253, but exhibit a strong horizontal connection of the diagonal maxima to the vertical axis, with "C" shape around the center. Therefore, the pair P07905-P07906 does not share features with the other paintings and are likely to have a fabric from the same origin, even if they have more than one thread of difference in the horizontal threads counting, where the standard deviation is much lower. Similar conclusions can be drawn for the pair P01182-P03253. The fabrics of paintings P01627 and P01819 exhibit a cross shape and even a high value along an axis, but different behavior around the center. Besides, recall that they have large differences in the thread counting. Therefore, they cannot be matched between them, nor to any other painting in Fig. 16.

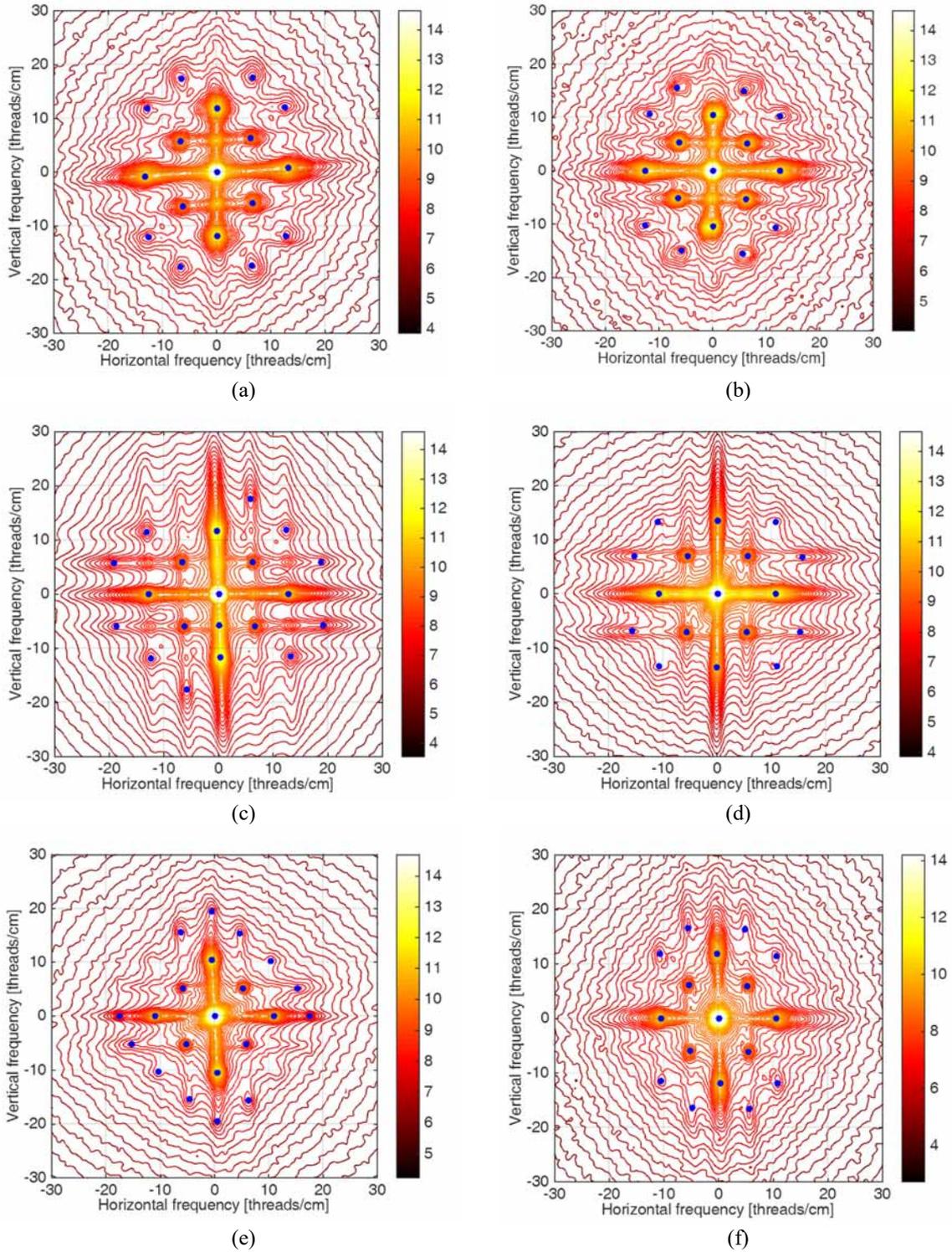

Fig. 16: PSD analysis of (a) P07905, (b) P07906, (c) P01627, (d) P01819, (e) P01182 and (f) P03253 for $N = 400$ Blackman-Harris window, $D = 100$ and $N_{FFT} = 2048$. Dots are included in places where theoretical maxima should be located and we found local maxima for the PSD nearby.

Table 4: PSD features results for several paintings of the 17th century.

|  | Shape | Diagonal | Centre | Axes |
|---|---|---|---|---|
| P07905 | Diamond | Horizontal | C | - |
| P07906 | Diamond | Horizontal | C | - |
| P01627 | Cross | - | C | Vertical |
| P01819 | Cross | - | O | Horizontal |
| P01182 | Diamond | - | \ | - |
| P03253 | Diamond | - | \ | - |

# 6. Conclusions

In this paper, we have proposed a new method to analyze and characterize the fabric of canvases. This technique is based on two different and complementary viewpoints. First, a theoretical analysis of the weave in the frequency domain and the use of triangles in it to find the main characteristics of the threads. Second, the use of the PSD rather than the DFT and the search of distinctive features in it. The result is a fingerprint that portray the whole canvas. We have shown that this approach is of great utility for both thread counting and paintings pairing, even in situations where previous methods fail, like very old and deteriorated artworks or simply canvases of small size. This information, combined with the study of historical sources or artistic and chemical analyses, help the museums in their research, such as dating or attribution of the paintings.

# Acknowledgments

The authors are grateful to Laura Alba, curator at the Museo Nacional del Prado, for her support, comments and suggestions on this work.